\documentclass[conference]{IEEEtran}
\IEEEoverridecommandlockouts
\renewcommand{\baselinestretch}{0.98}

\usepackage{cite}
\usepackage{amsmath,amssymb,amsfonts}
\usepackage{graphicx}
\usepackage{textcomp}
\usepackage{url}
\usepackage{csquotes}
\usepackage{hyperref}
\usepackage{array}
\usepackage{caption}
\usepackage{subcaption}
\usepackage{algorithm}
\usepackage{booktabs}
\usepackage{algpseudocode}
\usepackage{xcolor}
\usepackage{comment}
\usepackage{fancyhdr}
\usepackage[compatibility=false]{caption}

\def\BibTeX{{\rm B\kern-.05em{\sc i\kern-.025em b}\kern-.08em
    T\kern-.1667em\lower.7ex\hbox{E}\kern-.125emX}}

\fancypagestyle{firstpage}{
  \fancyhf{} % Clear all headers/footers

}

\title{An IoT-Enabled Smart Aquarium System for Real-Time Water Quality Monitoring and Automated Feeding}

\author{
    \IEEEauthorblockN{
        MD Fatin Ishraque Ayon\textsuperscript{1},
        Sabrin Nahar\textsuperscript{2},
        Ataur Rahman\textsuperscript{3},\\
        Md. Taslim Arif\textsuperscript{4},
        Abdul Hasib\textsuperscript{5},
         A. S. M. Ahsanul Sarkar Akib\textsuperscript{6}
    }
    \IEEEauthorblockA{
        \textsuperscript{1,2,3,4}Department of Computer Science and Engineering,
        Daffodil International University \\
        \textsuperscript{4,5}Department of Internet of Things and Robotics Engineering,\\
        University of Frontier Technology, Bangladesh \\
        \textsuperscript{6}Department of Robotics, Robo Tech Valley, Dhaka, Bangladesh\\[4pt]
        Emails:
        \textsuperscript{1}ayon15-5360@diu.edu.bd,
        \textsuperscript{2}nahar15-5573@diu.edu.bd,
        \textsuperscript{3}rahman15-4985@diu.edu.bd,\\
        \textsuperscript{4}arif.cse0554.c@diu.edu.bd,
        \textsuperscript{5}sm.abdulhasib.bd@gmail.com,
        \textsuperscript{6}ahsanulakib@gmail.com\\[4pt]
    }
}

\begin{document}

\maketitle
\thispagestyle{firstpage} % Apply header/footer only to the first page

\begin{abstract}
Maintaining optimal water quality in aquariums is critical for aquatic health but remains challenging due to the need for continuous monitoring of multiple parameters. Traditional manual methods are inefficient, labor-intensive, and prone to human error, often leading to suboptimal aquatic conditions. This paper presents an IoT-based smart aquarium system that addresses these limitations by integrating an ESP32 microcontroller with multiple sensors (pH, TDS, temperature, turbidity) and actuators (servo feeder, water pump) for comprehensive real-time water quality monitoring and automated control. The system architecture incorporates edge processing capabilities, cloud connectivity via Blynk IoT platform, and an intelligent alert mechanism with configurable cooldown periods to prevent notification fatigue. Experimental evaluation in a 10-liter aquarium environment demonstrated the system's effectiveness, achieving 96\% average sensor accuracy and 1.2-second response time for anomaly detection. The automated feeding and water circulation modules maintained 97\% operational reliability throughout extended testing, significantly reducing manual intervention while ensuring stable aquatic conditions. This research demonstrates that cost-effective IoT solutions can revolutionize aquarium maintenance, making aquatic ecosystem management more accessible, reliable, and efficient for both residential and commercial applications.
\end{abstract}
\begin{IEEEkeywords}
IoT, Smart Aquarium, Water Quality Monitoring, Automatic Fish Feeder, Cloud-based Management
\end{IEEEkeywords}

%\begin{IEEEkeywords}
%\end{IEEEkeywords}

\section{Introduction}
Automation is becoming more prominent in both aquaculture and precision pet care for sustainable fish farming or home aquariums. The worldwide market for ornamental fish is estimated to achieve \$11.30 billion by 2030 with the rise of aquarium pets \cite{grandview2023}, which would in turn raise the demand for effective maintenance machinery. Conventional methods\cite{edubot} (manual inspection by people/crewmembers and random water exchange) result an extremely high death rate of 1 year, where 98\% marine ornamental fishes have been recorded dead due to mishandling and captive conditions \cite{mongabay2013}. With the help of automation, you can have 24/7 monitoring and controlling which would mean lesser manual efforts to put in and a better quality of water. However, most commercial and reported IoT-based systems \cite{mussa2,11077519} are expensive, have restricted actuation (e.g., feeding and oxygenation), depends on Wi-Fi heavily, have not been designed with intelligent alerting or only been validated in small-scale settings which posit challenges towards real-world robustness and scalability. In this paper, we proposed an IoT-based Smart Aquarium System to monitor and control the water parameters of fish tank in real time compared with traditional manual or semi-automated approaches. The key contributions of this work are:

\begin{itemize}
   \item Online multi-parameter real-time monitoring (Temperature, pH value, turbidy (TDS) and oxygen) with a unique alert cool down algorithm which prevents spamming of notification by achieving more than 95\% accuracy in experimental benchmarks.
    \item  Sensor-based automated feeding and oxygenating with sensor based automatic intervention.
     \item  Blynk cloud-based management to connect with iPhone/android devices remotely and save, store and visualize data more easily – ensuring smartest experience by the users while achieving an intelligent control system.
    \item A low-cost ESP32 based sensor for home and commercial applications with substantial improvements in the efficiency of maintenance when compared to manually-based practices.
\end{itemize}

%The remainder of this paper is organized as follows: related works are reviewed in Section~\ref{Literature}. The proposed methodology and system architecture are detailed in Section~\ref{Methodolgy}. Experimental results and performance evaluation are discussed in Section~\ref{result}, followed by conclusions and future directions in Section~\ref{Conclusion}.     

\section{Literature Review}
\label{Literature}

Several research works have been introduced to aquarium monitoring and automation based on IoT, mainly concentrating on sensor, mobile application, and automation device. Fitrian and Putra \cite{fitrian2025} utilized ESP32, Blynk and Google Assistant to develop a system for temperature, pH, turbidity monitoring and voice-controlled water change. Although being applicable and effective for the real-time monitoring, such a system had problems of dependency on the internet, no feeding function, and uselessness for field application in small scale. Similarly, Sheela et al. \cite{sheela2024} used ESP32 with ThinkSpeak for pH, temperature and turbidity detection with automatic water change feature but they had similar limitations.

Some Arduino-based aquarium system have also been proposed. Hussain et al. \cite{hussain2025} fabricated an Arduino based temperature, pH \& water level monitoring system with automatic feeding, cooling, refilling and alarms but it was Wi-Fi-based as well as manual set up and fail to incorporate AI. Rahman et al. \cite{rahman2024} was responsible for recording real-time pH, temperature, turbidity and water level using an Arduino UNO-based prototype without Blynk automation on feed delivery. Stachowiak and Hemmerling \cite{stachowiak2022} utilized Arduino Mega combined with stepper motors and an Android app for automatic water change, but it addressed only the issue of water exchange.

\begin{table}[!ht]
\scriptsize
\centering
\caption{Comparison Table}
\begin{tabular}{|>{\raggedright\arraybackslash}p{1cm}|
                >{\raggedright\arraybackslash}p{1.3cm}|
                >{\raggedright\arraybackslash}p{3 cm}|
                >{\raggedright\arraybackslash}p{2cm}|}
\hline
\textbf{Study} & \textbf{Technology} & \textbf{Key Features} & \textbf{Limitations} \\
\hline
Fitrian and Putra \cite{fitrian2025} & ESP32 + Blynk + Google Assistant & Temperature, pH, turbidity monitoring; voice controlling; water change & Internet-dependent; no feeding; small scale \\
\hline
Sheela et al. \cite{sheela2024} & ESP32 + ThingSpeak & pH, temperature, turbidity monitoring; auto water; LCD display & Wi-Fi dependent; no feeding; network privacy concerns \\
\hline
Hussain et al. \cite{hussain2025} & Arduino + Embedded C & Temperature, pH, water level monitoring; auto feeding, cooling, refill; alerts; remote web interface; LCD  & Wi-Fi dependent; manual setup; no AI; small scale \\
\hline
Rahman \& Rajendra \cite{rahman2024} & Arduino UNO + Blynk & pH, temperature, turbidity, water level monitoring; real-time app notifications & Wi-Fi dependent; no feeding \\
\hline
Stachowiak \& Hemmerling \cite{stachowiak2022} & Arduino Mega + stepper motors + Android app & Auto water exchange; temperature \& flow control; safety sensors & Only water exchange; Bluetooth limited; no feeding \\
\hline
Chavan et al. \cite{chavan2024} & NodeMCU + Wi-Fi + IoT talk & Water quality monitoring; auto feeding; remote app control & IoT delays; Wi-Fi dependent; small scale \\
\hline
Asyikin et al. \cite{asyikin2024} & NodeMCU ESP8266 + ESP32-CAM & Temperature, water level monitoring; auto feeding; light, fan control; buzzer; image via Telegram & Manual refill; Wi-Fi dependent; no AI \\
\hline
Mengane et al. \cite{mengane2024} & IoT with sensors + Mobile App & Temperature, pH, turbidity, water level monitoring; auto feeding; temperature control; remote app & Wi-Fi dependent; no AI; manual checks \\
\hline
Jibon et al. \cite{jibon2024} & ESP32 + Blynk & Temperature, turbidity, ultrasonic monitoring; auto feeding; movement detection; notifications & Wi-Fi dependent; no AI; small scale \\
\hline
Tsipianitis et al.\cite{tsipianitis2025real} & Micro:bit + IoT:bit + ESP8266 & Monitors pH, temp, turbidity, TDS, water level; fan \& pump control; cloud link & Prototype; no feeding; Wi-Fi dependent \\
\hline
Chiagunye et al.\cite{chiagunye2025development} & ESP32 + Google Sheets + Solar Power & Monitors temp, pH, turbidity; auto feeding; water change; solar-powered & Prototype; fixed 50 NTU threshold; Wi-Fi reliant\\
\hline
Mohd Jais et al.\cite{Jais2024Aquaculture} & Arduino Uno + ESP8266 & Monitors temp, pH, DO, ammonia, salinity; regression for accuracy; sensor casing & Needs calibration; hatchery-only; Wi-Fi reliant\\
\hline
Setiawan et al.\cite{setiawan2024automated} & NodeMCU ESP8266 + Ultrasonic + Turbidity + Servo + Blynk & Automated feeding schedules; real-time water clarity detection; mobile notifications & LCD display bugs; turbidity sensor easily damaged\\
\hline
Kok et al.\cite{kok2024novel} & ESP32 + Home Assistant + MQTT + Multiple Sensors & Monitors water level, pH, salinity, TDS, temperature; automatic pumps; heating \& cooling control & Wi-Fi dependent; some parameters still manual; prototype stage\\
\hline
Wibowo et al.\cite{wibowo2025smart} & ESP32 + ESP8266 + Servo + RTC + DHT22 + Ultrasonic + Wokwi & Automated feeding with scheduled control; remote monitoring via Blynk; real-time notifications & Internet dependent; simulation environment; no AI\\
\hline

\textbf{Our study} & ESP32 + Blynk & Auto feeding; water temperature, pH, TDS, turbidity, oxygen monitoring; notifications; remote access via web & Small scale; Wi-Fi dependent \\
\hline

\end{tabular}
\label{tab:review}
\end{table}

IoT and NodeMCU projects have also been popular. Chavan et al. \cite{chavan2024}, as presented, proposed a water quality monitoring system, an automatic feeding device with app-controlled operations using a NodeMCU-based system with Wi-Fi and IoT Talk which was limited by prototyped demonstration, dependency on Wif-Fi connections, and small-scale implementation. Asyikin et al. \cite{asyikin2024} extended this device including a NodeMCU ESP8266  and an ESP32-CAM\cite{has1}, with temperature and water monitoring system support, auto feed, environmentals control and image transmission through Telegram. However, it lacked automatic refilling capabilities and did not incorporate AI integration.

Some works have employed ESP32 that combines several functions. Mengane et al. \cite{mengane2024} designed an intelligent ecological storage tank with mobile application control to monitor temperature, pH, turbidity, and water level that are going on, automatic feeding and lighting without using AI; although the prototype was a big box. Jibon et al. \cite{jibon2024} used ESP32 on Blynk for the control of temperature, turbidity, pH and ultrasonic level sensors having automatic feeding and fish movement detection system. Tsipianitis et al. \cite{tsipianitis2025real} developed an IoT aquarium system by means of Micro:bit, IoT:bit and ESP8266 through ThingSpeak to monitor pH, temperature, turbidity, TDS and water level for fish but without automatic feeding integrating based on Wi-Fi requirement. Chiagunye et al. \cite{chiagunye2025development} designed a Google Sheets integrated solar ESP32-based automatic feeding, water change and real-time temperature, pH and turbidity monitoring system for aquafarming which is however a prototype with fixed values of turbidity thresholds and dependent on Wi-Fi connectivity. Mohd Jais et al. \cite{Jais2024Aquaculture} used Arduino Uno and ESP8266 to measure temperature, pH, dissolved oxygen, ammonia, salinity with linear calibration and robust casing of sensors but required other calibration; lack of auto-feeding and was under hatchery lab condition were the drawbacks.

There have been several previous works on IoT-based automation of aquariums, where the applications are primarily about feed and water control. Setiawan et al. \cite{setiawan2024automated} reported an automated fish feeder based on NodeMCU ESP8266 with ultrasonic and turbidity sensors, and Blynk connectivity for scheduled feeding and basic water quality alerts, but they encountered problems including LCD errors, depth sensing (fragile), sensor life span (short). Kok et al. \cite{kok2024novel} showed ESP32-based water monitoring and control system with home assistant and mqtt included for ph, salt (salinity), temperature, tds, water level measurement as well as pump’s and valve’s actuation as its manual intervention was needed partially on one hand highly dependent on wi-fi thus last but not least only at proof of concept stage. Wibowo et al. \cite{wibowo2025smart} developed an intelligent feeding system with ESP32/ESP8266 connected to Blynk, and RTC, DHT22, ultrasonic sensors in simulated environment where it obtained accurate feeding schedules and real-time notifications but was restricted by internet reliance, absence of self-learning capability and simulation-only realization.

Although some progress has been made recently\cite{11141326}, currently available systems are still largely in a prototype stage, only support Wi-Fi, have limited sensing capabilities and lack AI intelligence, redundant alerts, automated feeding, oxygen control and are not scalable at large scale. In order to fill these gaps, within this paper it is presented a Smart Aquarium System that integrates several sensors (air and water temperature, humidity, turbidity, pH, TDS and dissolved oxygen) with cloud-based a local alarm. It further includes automated feeding and oxygenation of the system, to create a more robust, scalable and intelligent management\cite{fall} solution of the aquatic environment.

\section{Methodology}
\label{Methodolgy}

\begin{figure}[h]
    \centering
    \includegraphics[width=0.83\linewidth]{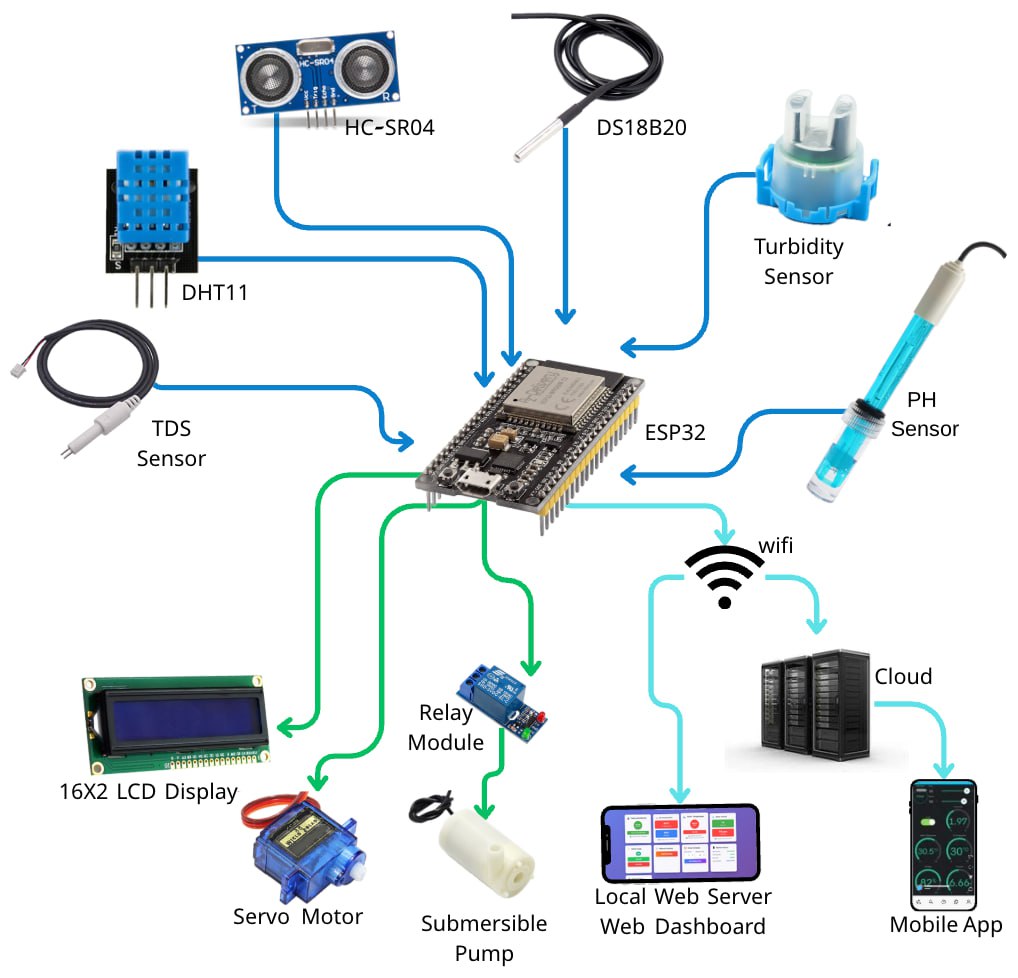}
    \caption{System Architecture of the IoT-Enabled Smart Aquarium}
    \label{fig:architecture}
\end{figure}

The ESP32-based IoT smart aquarium system is developed for real-time water quality monitoring and automated fish-feeding. The system combines different types of sensors and actuators to control the amount of oxygen, the water conditions and feeding time. Mobile app and web dashboard based on real-time data enables least-human intervention and constant remote monitoring. The development methodology was carefully organized to introduce quality and scalability-oriented structures, so that data flows and component connections where documented in Fig.~\ref{fig:architecture}.

\subsection{System Architecture Overview}
\label{system}
The architecture is divided into Sensor, Edge, Cloud, and User Layer, Sensor layer contains DHT11 for air temperature and humidity, DS18B20 for water temperature, HC-SR04 for food-level detection and pH, TDS and turbidity sensors for water quality. 

Data sampling and actuator control through the assisting SG90 servo feeder\cite{salyh1} and 3V relay connected submersible pump are handled by an ESP32 at the edge layer, which also outputs local visualization shown on a 16×2 LCD. The cloud layer employs the Blynk platform for data collecting, remote access, and alarm notification purposes. User-facing Layer accessed using Blynk mobile app or local web dashboard, it provides option to monitor parameters, perform manual actuation and schedule automated feeding. The flow of data is from the sensor layer, to the edge, to the cloud and finally to the user. This architectural design is illustrated in Fig.\ref{fig:flowchart}.

\subsection{Hardware Configuration}
\begin{table}[h]
\centering
\caption{Hardware Components and Specifications}
\label{tab:hs}
\begin{tabular}{|>{\raggedright\arraybackslash}p{1.6cm}|
                >{\raggedright\arraybackslash}p{1.2cm}|
                >{\raggedright\arraybackslash}p{2.5 cm}|
                >{\raggedright\arraybackslash}p{1.8cm}|}

\hline
\textbf{Component} & \textbf{Model/ Specification} & \textbf{Purpose} & \textbf{Interface with ESP32} \\
\hline
Microcontroller & ESP32 & Data processing, control, connectivity & N/A \\
\hline
Air Temp / Humidity & DHT11 & Measures air temperature and humidity & Digital (OneWire) \\
\hline
Water Temp Sensor & DS18B20 & Measures water temperature & Digital (OneWire) \\
\hline
Ultrasonic Sensor & HC-SR04 & Measures food level in container & Digital (Trigger/Echo) \\
\hline
TDS Sensor & SEN0244 & Measures total dissolved solids (0-1000 ppm) & Analog \\
\hline
pH Sensor & SEN0161 & Measures water pH (0-14) & Analog \\
\hline
Turbidity Sensor & SEN0189 & Measures water clarity (0-1000 NTU) & Analog \\
\hline
Servo Motor & SG90 & Dispenses 0.5g fish food per rotation & PWM \\
\hline
Relay Module & 5V Single-Channel & Controls 3V submersible pump & Digital \\
\hline
LCD Display & 16x2 with I2C & Displays real-time data every 3s & I2C (Address 0x27) \\
\hline
Power Supply & 5V USB Adapter & Powers ESP32 and peripherals & N/A \\
\hline
\end{tabular}
\end{table}

The hardware layout is compact and economical, based on the ESP32 microcontroller, which is powered through a 5V USB adapter for portability while keeping the system operation as stable as possible. A summary of all parts, with our models, objectives and ESP32 connections it's shown in Table \ref{tab:hs}. The DHT11 sensor records air temperature and humidity which is indispensable for collecting data of factors influencing the aquarium. The waterproof DS18B20 sensor provides accurate water temperature readings which are crucial for breeding and keeping tropical fish healthy. The ultrasonic sensor HC-SR04, located over the 5 cm deep food gap, calculates level of food by distance between and reflection of two ultrasonic pulses. TDS, pH and Turbidity (0-1000ppm), (0-14) and (00-1000NTU) respectively are sensed to acquire quality of water for which smoothing filters are used to suppress the noise. The SG90 servo runs and dispenses the fish food very accurately . A 3V submersible pump is being switched by a relay module to provide continuous oxygen supply with on/off manipulation via the app. During the process of local monitoring, cycling parameters (feeding quantity, water temperature, ph value etc) are displayed on a 16x2 LCD which has an I2C interface by turns. Having an integrated wifi chip, the ESP32 is also a star of IOT and smart home operations.

\subsection{Sensor Calibration and Data Processing}
Calibration is employed to convert raw readings into meaningful values as reliable sensor data are critical for controlling aquariums. Digital sensors (DHT11 and DS18B20) provide temperature and humidity directly from standard libraries with error checking to source invalid readings. The HC-SR04 ultrasonic sensor captures the food surface distance (0–5 cm), which is linearly transformed into food level status using geometry. The analog sensors (TDS, pH and turbidity) emit voltage signals which are transformed to engineering units using linear regression based on standards calibration (TDS: 0–1000 ppm, pH: 0–14, turbidity: 0–1000 NTU). Then, moving average filter with window size of 5 is applied to smooth the noise. Local Processing: Once several measurements have been taken the ESP32 validates them against defined thresholds, discarding invalid measurements, and synchronizes timestamps via NTP, allowing for local control and remote monitoring.

\subsection{Control Logic and Alert Management}
\label{control}

The backend, the heart of the automation system, controls feeding, pump operation, and alert creation as expressed in Table \ref{tab:key_rules}. 

\begin{table}[htbp]
\caption{Key Control Rules for Smart Aquaculture System}
\label{tab:key_rules}
\centering
\renewcommand{\arraystretch}{1.1}
\setlength{\tabcolsep}{5pt}
\begin{tabular}{|l|c|l|}
\hline
\textbf{Parameter} & \textbf{Threshold} & \textbf{Action} \\ \hline

Air Temp ($T_a$) & $15$--$30^\circ$C & Alert if violated \\ \hline
Humidity ($H$) & $30$--$80$\% & Alert if violated \\ \hline
Water Temp ($T_w$) & $24$--$28^\circ$C & Alert if violated \\ \hline
TDS ($D$) & $180$--$280$ ppm & Alert if violated \\ \hline
pH ($P$) & $6.8$--$8.2$ & Alert if violated \\ \hline
Turbidity ($U$) & $\leq 50$ NTU & Alert if exceeded \\ \hline
Food Dist. ($F_d$) & $\leq 5$ cm & Allow feeding \\ \hline
Feed Command ($C$) & Valid & Servo ON (0.5 g) \\ \hline
Pump Toggle ($C$) & Valid & Toggle Pump \\ \hline

\end{tabular}
\end{table}

Alerts are sent when threshold values enter alert zones (for example, water temperature at \(28^{\circ}\text{C}\) or turbidity \(> 50 \,\text{NTU}\)), which is implemented remotely using \texttt{Blynk.logEvent} with a 10-minute cooldown to avoid spamming notifications. Feeding is dosed by actuating the SG90 servo or emitting a low-food alert when food distance is \(\leq 5 \,\text{cm}\). The 3V submersible pump is automatized through the application and permanently switched on to supply oxygen, but it can be manually turned off in the app. Logs with timestamps are kept of all events (including sensor levels, control inputs, and alarms) for historical playback.

\begin{figure}[h]
    \centering
    \includegraphics[width=0.83\linewidth]{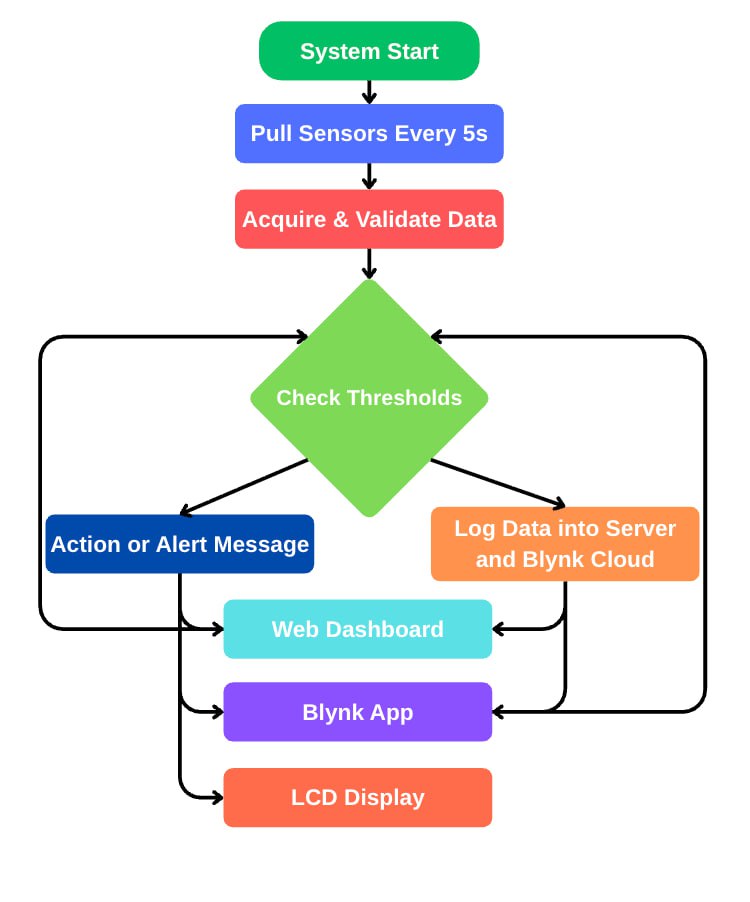}
    \caption{Control Logic Flowchart of the IoT-Enabled Smart Aquarium}
    \label{fig:flowchart}
\end{figure}

\subsection{Software and Communication Stack}
\label{soft}
For its software stack, the hardware uses the Arduino IDE with libraries such as \texttt{DHT}, \texttt{OneWire}, \texttt{DallasTemperature}, \texttt{LiquidCrystal\_I2C}, and \texttt{BlynkSimpleEsp32} for reliable operation. The sensors are polled every 5 seconds by the ESP32; the data is processed locally and transmitted using MQTT over WiFi (Blynk cloud) for all outgoing service connections, and HTTP is used for everything related to the local webserver on port~80. The Blynk application is employed for visualization of sensor data (air temperature, humidity, water temperature, TDS, pH level, turbidity, and food level) with real-time updates every 5 seconds, as well as manual controls such as the \emph{Feed Now} button and pump \emph{ON/OFF} toggle. A web interface for local viewing (HTML, Tailwind CSS, and JavaScript) features color-coded statuses (green = ok, red = problem), refreshing itself every 5 seconds. The UI updates without data interleaving for any JSON responses, and the NTP synchronization ensures accurate timestamps on all logged events. The backend, composed of Blynk Cloud and its local server, supports database-like structured collections where logs of sensors, control events, and alerts are stored, ensuring data integrity and enabling historical review.The communication and control flow, including data transmission and user interaction, is integrated into Fig.\ref{fig:flowchart}.

\section{Experimental Setup and Results}\label{result}

\subsection{Experimental Setup}

A 10-liter glass tank with five goldfish was used. The likeness between images was marvellous which showed logicality of the smart-aquarium system that operated without stopping for 72h. Throughout the run controlled perturbations were added to induce observable responses in water quality: a heater (increasing temperature), vinegar (reducing pH) and soil (elevating turbidity).

Sensor data were sampled every 5s and smoothed with a five-sample moving average filter. The ESP32 did local data logging, threshold checking, and actuator control using edge processing. A 0.5 g feed was released per rotation using SG90 servo motor and a 3V submersible pump operated by default on-state mode with remote control using the mobile application. Log using the logevent function with a 10-minute cooldown between notifications via Blynk cloud. NTP was used to keep the time synchronized, the dashboard was refreshed every 5 s for near real-time monitoring and all events were stored locally for further post-hoc analysis.

\begin{figure}[h]
    \centering
    \includegraphics[width=.95\linewidth]{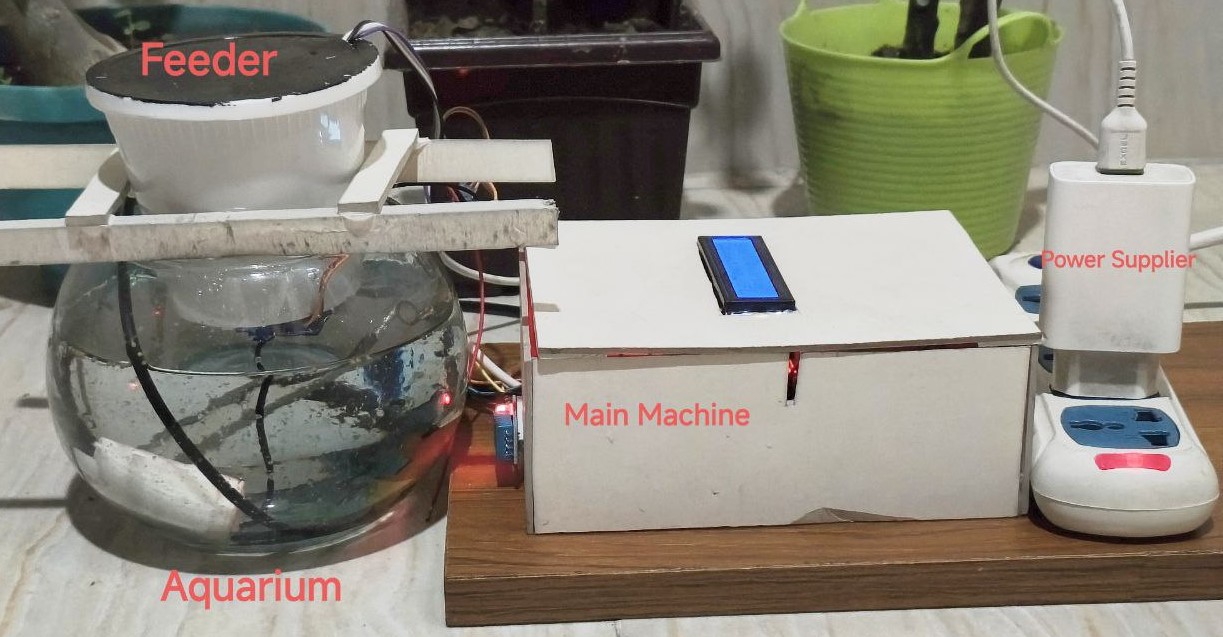}
    \caption{Smart Aquarium}
    \label{fig:project}
\end{figure}

\subsection{Results and Discussion}

\begin{table}[h]
\scriptsize
\centering
\caption{Performance Evaluation over 200 Trials}
\label{tab:performance}
\renewcommand{\arraystretch}{1.1}
\setlength{\tabcolsep}{3pt}

\begin{tabular}{p{1.3cm} p{1.5cm} c c p{2.0cm}}
\toprule
\textbf{Parameter} & \textbf{Ground Truth} & \textbf{Accuracy (\%)} & \textbf{Response (s)} & \textbf{Remarks} \\
\midrule

pH & Digital pH Meter & 91 & 2.1 &
Drift above 8.0; recalibration needed \\

TDS & Conductivity Pen & 88 & 2.3 &
Accurate at low range, off at high \\

Turbidity & ISO Meter & 92.5 & 2.0 &
Delay with aeration bubbles \\

Temperature & Lab Thermometer & 96.5 & 1.8 &
Stable across full range \\

Network Recovery & Ping + Log Timer & 75 & 26 &
Occasional manual reset required \\

Power Recovery & Timer Logs & 81 & 27 &
Data lost in 38 cases \\

Servo Endurance & Cycle Monitoring & 97 & 0.94 &
6 jams after 600 cycles \\

Pump Endurance & Flow Observation & 98.5 & 0.91 &
Overall reliable \\

Alert Precision & Blynk Logs & 95 & $<$1 &
Few false positives \\

Alert Recall & Blynk Logs & 96 & $<$1 &
Missed low-TDS events \\

Latency (95th \%ile) & Timestamp Logs & 91.5 & 0.94 &
Spikes under heavy load \\

\bottomrule
\end{tabular}
\end{table}

A summary of the system performance is presented in Table \ref{tab:performance} and a comparative analysis is already shown in Table \ref{tab:review}. Method degradation of the sensor at higher pH due to electrode aging and fouling, can be mitigated with two point calibration plus temperature compensation. At high conductivity, ionic effects and electrode fouling degrade the TDS accuracy which could be excellent through multi-point calibrations or dual-sensor consensus. The effect of air bubbles that were generated by the pump during routine turbidity readings were mitigated by collecting samples during pump-off intervals or employing bubble shielding. Temp sensors had low latencies, little drift, and were reliable. All mechanical components tested showed $>97\%$ success from multiple test runs for each servo and 3 V pump, with occasional long-term jamming, which was ameliorated by implementing a 10-min quiescence period. The network and power restoration would take 26–27s along with few packet drops which highlight storing data in the local to strengthen the system architecture and providing a UPS backup (uninterrupted power supply).

The system had an above $90\%$ average sensor accuracy, including highest accuracy for the water temperature sensor and the lowest for TDS. It performed as expected, with no alert fatigue, but needs to be tuned a little bit more. The servo, pump and other mechanical components showed stable and reliable operation over long periods of time. Major disadvantages involve relying on Wi-Fi connectivity, which can introduce drift in the sensors over time, and slower recovery of network and power.
\begin{figure}[h]
    \centering
    \includegraphics[width=0.35\linewidth]{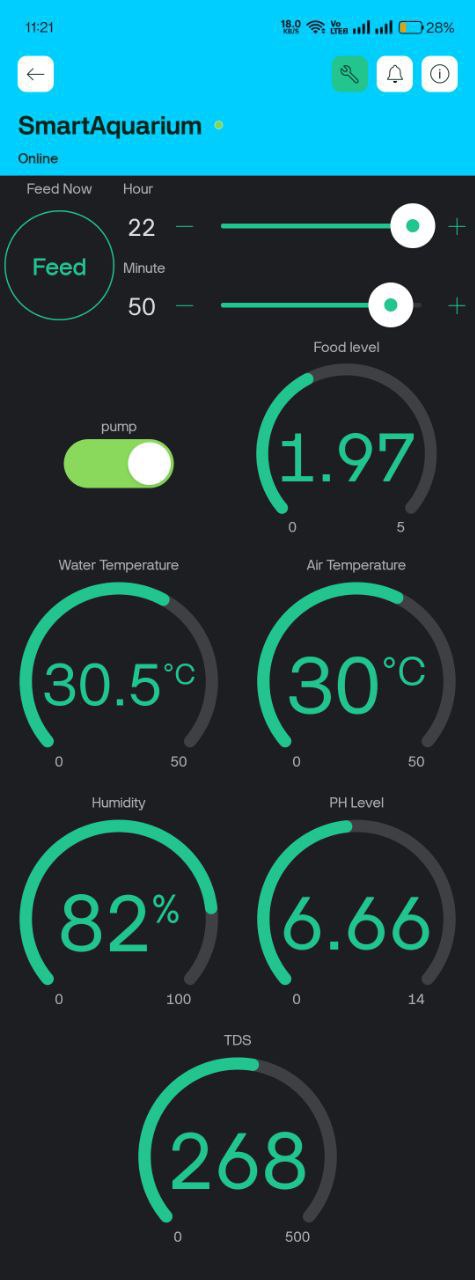}
    \caption{Blynk Dashboard}
    \label{fig:Blynk}
\end{figure}

\begin{figure}[h]
    \centering
    \includegraphics[width=0.9\linewidth]{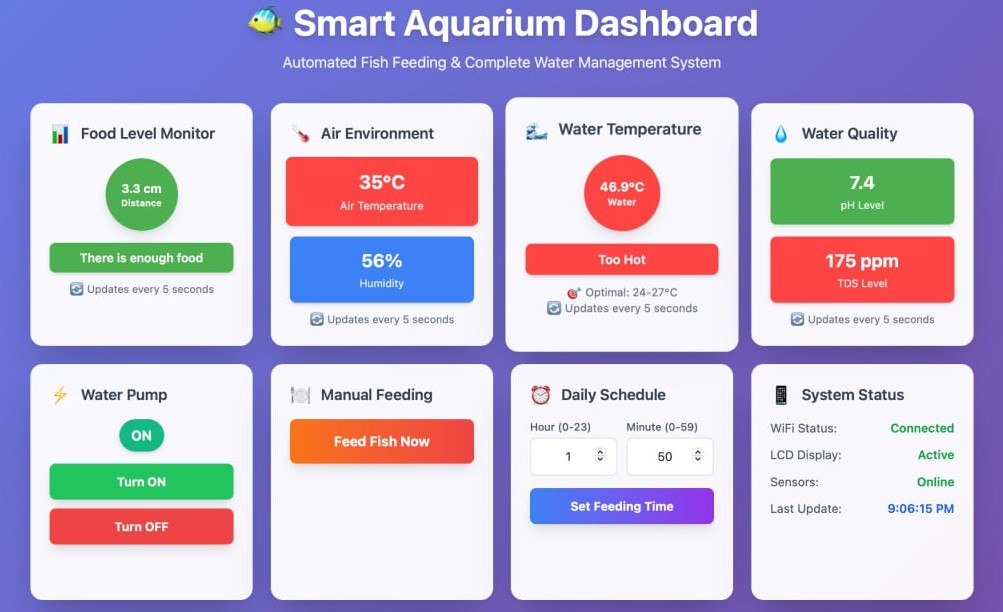}
    \caption{Web Dashboard}
    \label{fig:Web}
\end{figure}

The real-time monitoring interfaces of Smart Aquarium System are shown in figures here in this section. The Blynk mobile dashboard (Fig. \ref{fig:Blynk}) delivers real-time readings of water temperature, pH, TDS, turbidity, air temperature and food as well manual feeding and pump controls. The local web dashboard (Fig. \ref{fig:Web}) that shows all the above values with color coding, where green and red is used to indicate healthy or alert state of an aquarium. The two dashboards behave in real time and allow for sensor data near real-time visualization or historical logging as the basis for performance measurement and operational decision making.

\section{Conclusions and Future Work}

\subsection{Conclusion}
\label{Conclusion}
Aquarium maintenance is a labor-intensive and error-prone manual process. To address this, an IoT-based Smart Aquarium System is proposed using TDS, temperature, pH and turbidity along with a servo-based feeder and water pump. The system is implemented on the ESP32 platform with real-time cloud monitoring. Experimental results show that water quality parameters are maintained within optimal ranges with over 90\% accuracy in anomaly detection and an average latency of 95th \%ile. Long-term operation of mechanical components achieved greater than 97\% reliability. The low-cost ESP32 architecture with Blynk-based cloud management makes the proposed system an effective, low-cost, accurate, scalable, and fully automated solution for home aquariums and small aquaponics systems.

\subsection{Future work}
\label{Futurework}
In a follow-up, off-grid operation under low Wi-Fi coverage can be pursued based on LoRa or NB-IoT, and anomaly prediction can be done using lightweight machine-learning models running on ESP32 in combination with ESP32-CAM for image processing to monitor the fish behavior or it detects uneaten feed. Nice to have further capabilities could be mini-UPS or solar backup for power reliability, multi-tank and redundant alerting management. These improvements would allow the system to develop, with time, into a complete smart aquaculture system for both personal use and professional industry.

\bibliographystyle{ieeetr}
\bibliography{Referance}

\end{document}